\documentclass{article} 
\usepackage{iclr2023_conference,times}

\usepackage{amsmath,amsfonts,bm}

\def\eqref#1{equation~\ref{#1}}

\def\1{\bm{1}}

\DeclareMathAlphabet{\mathsfit}{\encodingdefault}{\sfdefault}{m}{sl}
\SetMathAlphabet{\mathsfit}{bold}{\encodingdefault}{\sfdefault}{bx}{n}

\usepackage{hyperref}
\usepackage{url}
\usepackage{graphicx}
\usepackage{wrapfig}
\usepackage{float}
\usepackage{caption}

\title{Mark My Words: Dangers of Watermarked\\Images in ImageNet}

\iclrfinalcopy

\author{
Kirill Bykov\textsuperscript{1, 2, *} \& Klaus-Robert M\"uller\textsuperscript{1, 3, 4, 5} \& Marina M.-C. Höhne \textsuperscript{1, 2, 3, 6, 7} \\
\textsuperscript{1}Technische Universität Berlin, Machine Learning Group, 10587 Berlin, Germany\\
\textsuperscript{2}Understandable Machine Intelligence Lab, ATB, 14469 Potsdam, Germany\\
\textsuperscript{3}BIFOLD -- Berlin Institute for the Foundations of Learning and Data, 10587 Berlin, Germany\\
\textsuperscript{4}Korea University, Department of Artificial Intelligence, Seoul 136-713, Korea \\
\textsuperscript{5}Max Planck Institute for Informatics, 66123 Saarbrücken, Germany \\
\textsuperscript{6}Machine Learning Group, UiT the Arctic University of Norway, 9037 Tromsø, Norway\\
\textsuperscript{7}Department of Computer Science, University of Potsdam, 14476 Potsdam, Germany\\
\textsuperscript{*}Corresponding Author: \texttt{KBykov@atb-potsdam.de}\\
}

\begin{document}

\maketitle

\begin{abstract}
The utilization of pre-trained networks, especially those trained on ImageNet, has become a common practice in Computer Vision. However, prior research has indicated that a significant number of images in the ImageNet dataset contain watermarks, making pre-trained networks susceptible to learning artifacts such as watermark patterns within their latent spaces. In this paper, we aim to assess the extent to which popular pre-trained architectures display such behavior and to determine which classes are most affected. Additionally, we examine the impact of watermarks on the extracted features. Contrary to the popular belief that the Chinese logographic watermarks impact the ``carton'' class only, our analysis reveals that a variety of ImageNet classes, such as ``monitor'', ``broom'', ``apron'' and ``safe'' rely on spurious correlations. 
Finally, we propose a simple approach to mitigate this issue in fine-tuned networks by ignoring the encodings from the feature-extractor layer of ImageNet pre-trained networks that are most susceptible to watermark imprints.

\end{abstract}

\section{Introduction}

In recent years, the utilization of ImageNet \cite{deng2009imagenet} pre-trained models has become a standard practice in Computer Vision applications \cite{kornblith2019better}. Trained on the large and diverse collection of images, these models obtain the ability to extract high-level visual features that later could be transferred to a different task. This technique, referred to as transfer learning (see e.g.~\cite{weiss2016survey} for a review), has proven to be highly effective, leading to significant advancements in various computer vision applications, such as object detection \cite{talukdar2018transfer}, semantic segmentation \cite{van2018transfer} and classification \cite{yuan2021large}.

Deep Neural Networks (DNNs), despite being highly effective across a variety of applications, are prone to learning spurious correlations, i.e., erroneous relationships between variables that seem to be associated based on a given dataset but in reality lack a causal relationship \cite{izmailov2022spurious}. This phenomenon, referred to as the ``Clever-Hans effect'' \cite{Lapuschkin2019ch} or ``shortcut-learning'' \cite{geirhos2020shortcut}, impairs the model's ability to generalize. In Computer Vision (CV), such correlations may manifest as DNNs' dependence on background information for image classification \cite{xiao2020noise}, textural information \cite{geirhos2018imagenet}, secondary objects \cite{rosenfeld2018elephant}, or unintended artifacts, such as human pen markings in skin cancer detection \cite{anders2022finding} and patient information in X-ray images for pneumonia detection \cite{zech2018variable}.

Recent studies have uncovered the presence of spurious correlations in the ImageNet dataset, specifically, the connection of the Chinese logographic watermarks to the ``carton'' class \cite{anders2022finding, li2022Dilemma}. These correlations make ImageNet-trained networks vulnerable to learn watermark detectors in their latent space, leading to incorrect predictions when encountering similar patterns in the data. Furthermore, it has been shown that this behavior persists even after fine-tuning on different datasets \cite{bykov2022dora}, indicating that the vulnerability to watermarks is not exclusive to ImageNet networks but possibly extends to all fine-tuned models.

With this study, we aim to examine which specific ImageNet classes are influenced by the artifactual behavior of watermarks. We analyze the extent to which commonly used pre-trained architectures exhibit this phenomenon and propose a straightforward solution for reducing such behavior in transfer learning by eliminating the most artifact-sensitive representations, with negligible effect on the model's performance.

\section{Method}


\begin{wrapfigure}{r}{0.45\textwidth}
\vspace{-0.45cm}
\centering
\includegraphics[width=0.45\textwidth]{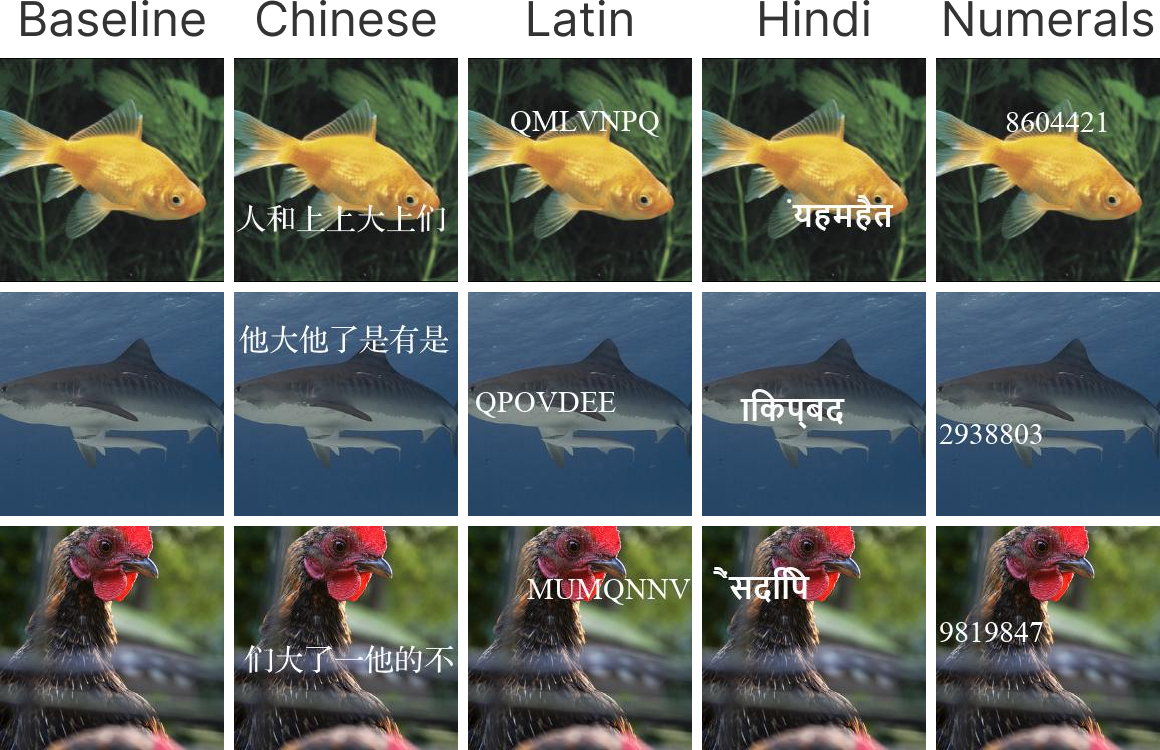}
\vspace{-0.5cm}
\caption{The illustration shows the images in the baseline dataset and their corresponding watermarked versions.}
\vspace{-0.25cm}
\label{fig:dataset_example}
\end{wrapfigure}

In this work, we define neural representations as sub-functions of a model that map the input domain to a scalar value indicating the activation of a specific neuron. Our analysis focuses on two primary scenarios: scalar representations of output classes and \textit{feature-extractor} representations, which correspond to the layer preceding the output logit layer\footnote{In the case of neurons that produce multi-dimensional activations in \textit{feature-extractor} representations, such as convolutional neurons, the channel neurons were analyzed by taking the average of the activation maps per each channel.}. 

To evaluate the susceptibility of individual representations to watermarks, we created binary classification datasets between normal and watermarked images and assessed their ability to distinguish between the two classes. We followed the approach outlined in \citet{bykov2022dora} and used a baseline dataset of 998 ImageNet images\footnote{Images were obtained from \url{https://github.com/EliSchwartz/imagenet-sample-images}, excluding 2 images that already contained Chinese logographic watermarks.}. We created four probing datasets by inserting random textual watermarks in the three most popular languages (Chinese, Latin, Hindi) \cite{sanches2014community} and Arabic numerals, as illustrated in Figure \ref{fig:dataset_example}. We evaluated the representations' ability to differentiate between watermarked and normal classes using AUC ROC, a widely used performance metric for binary classifiers. To do so, we utilized the true labels provided by the two datasets, where class 1 represents images with a watermark and class 0 represents those without. We first calculated the scalar activations from a specific neural representation for all images from both classes. Then, utilizing the binary labels, we calculated the AUC ROC classification score based on the differences in activations.
AUC ROC score of 1 indicates a perfect classifier, ranking the watermarked images consistently higher than normal ones, and 0.5 a random classifier. However, we can also observe scores less than 0.5, such as the score of 0 illustrating the perfect classifier, that is de-activated by the watermarked images. To measure the general ability of representations to differentiate between the two classes and provide evidence that the concept has been learned, we defined a \textit{differentiability} measure $d = \max\left(A, 1 - A\right), $ where $A$ is the AUC ROC score of the representation in the particular binary classification problem.

\section{Results}

To analyze the effects of watermarked images on learned representations, we employed 20 popular ImageNet-pre-trained Computer Vision architectures, namely AlexNet \cite{krizhevsky2014one}, ResNet 18, 50, 101, and 152 \cite{he2016deep}, ResNext 101 \cite{xie2017aggregated}, WideResNet 101 \cite{zagoruyko2016wide}, ViT \cite{dosovitskiy2020image}, BEiT \cite{bao2021beit}, Inception V3 \cite{szegedy2016rethinking}, DenseNet 121, 161, and 201 \cite{huang2017densely}, GoogLeNet \cite{szegedy2015going}, MobileNet V2 \cite{sandler2018mobilenetv2}, ShuffleNet V2 \cite{ma2018shufflenet},  VGG 11, 13, 16, and 19 \cite{simonyan2014very}.

\begin{figure}[h]
\begin{center}
\includegraphics[width=\textwidth]{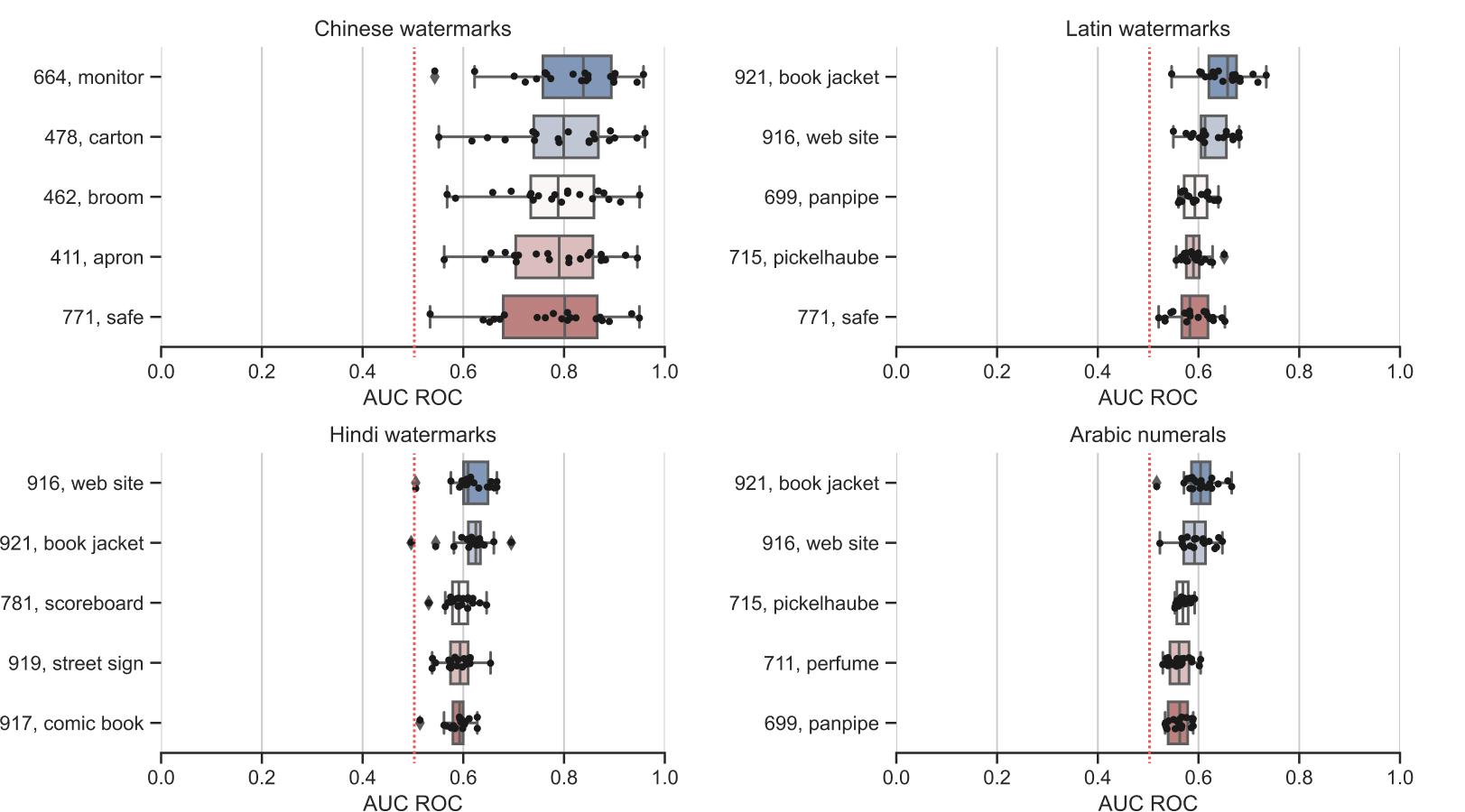}
\end{center}
\caption{ImageNet classes with the highest mean AUC ROC scores across the models analyzed in 4 different scenarios (Chinese, Latin, Hindi, and Numeric watermarks). Each dot represents the AUC ROC performance of the class representation for a single model.}
\vspace{-0.3cm}
\label{fig:imagenet_classes_auc_roc}
\end{figure}

For the 4 different scenarios, we collected the AUC ROC scores for every class logit representation across all 20 ImageNet pre-trained networks. Figure \ref{fig:imagenet_classes_auc_roc} illustrates the top-5 ImageNet classes by the highest average AUC ROC across the 20 models. We can observe the clear distinction between the different scenarios -- Chinese watermarks show significantly higher average classification scores, compared to the other three watermarks, namely Latin, Hindi, and Arabic numerals. Furthermore, it can be observed that classes with a high capability for detecting Chinese watermarks are not inherently linked to textual objects, whereas classes for other watermarks have a natural association with text, such as ``web site'' or ``book jacket''. This observation supports the conclusion that the ability of DNNs to detect Chinese logograms results from the Clever-Hans effect and is not desirable, whereas this cannot be said for other text detectors. Interestingly, by analyzing the classes with the lowest average AUC ROC we could even reveal --- for the first time --- the ability of ImageNet classes to detect the absence of the Chinese watermarks in images, which was not given for the other types of watermarks (illustrated in the Appendix \ref{fig:appndx:imagenet_classes_auc_roc_lowest}).

\begin{figure}[h]
\begin{center}
\includegraphics[width=\textwidth]{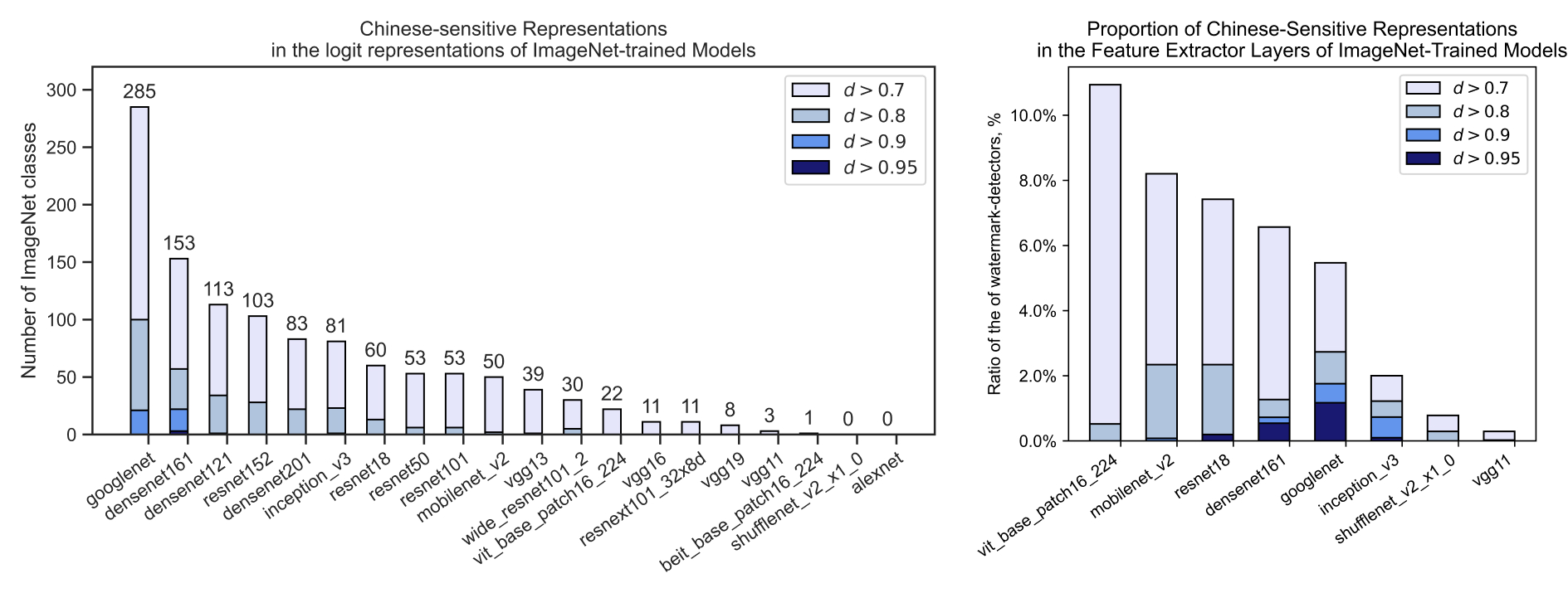}
\end{center}
\vspace{-0.5cm}
\caption{\textit{Left}: Number of output class representations that exhibit a high degree of differentiability towards Chinese watermarks across various ImageNet models. \textit{Right}: Percentage of representations in the feature-extractor layers of various networks that demonstrate a high degree of differentiability towards Chinese watermarks.}
\vspace{-0.3cm}
\label{fig:imagenet_classes_barplot}
\end{figure}

Figure \ref{fig:imagenet_classes_barplot} illustrates the number of representations that are sensitive to the Chinese symbols, across the logit and feature-extractor layers (layers of representations, preceding the last prediction layer) of different networks. From the left figure, which represents the sensitivity of output logits, we can observe that nearly all of the networks exhibit sensitive logit representations. This could be the reason for the average drop of 10.6\% in model performance when transparent Chinese watermarks are added to the ImageNet validation dataset, as reported in \cite{li2022Dilemma}. Some networks, such as GoogleNet, have up to 285 output classes (out of 1000) that are susceptible to Chinese watermarks. The right figure, which represents the ratio of sensitive representations to the total number of representations in the feature-extractor layers, reveals a significant proportion of representations that have a high degree of differentiability toward the Chinese watermarks. Furthermore, we can observe that several networks, including DenseNet-161, ResNet-18, and GoogLeNet exhibit at least several representations with very high watermark differentiability scores ($d > 0.95$), which is in line with the reported high number of Chinese-sensitive class representations across output logit layer.

\section{Ignoring sensitive embeddings during fine-tuning}

\begin{figure}[h]
\begin{center}
\includegraphics[width=\textwidth]{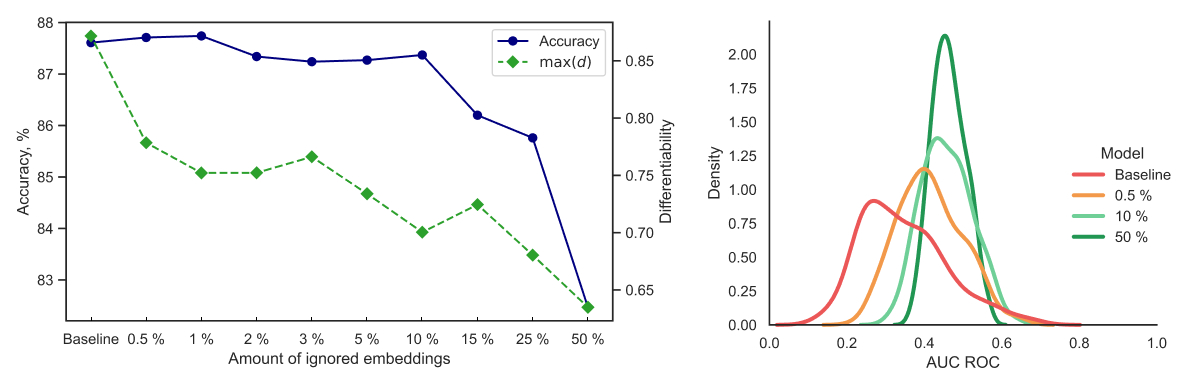}
\end{center}
\vspace{-0.4cm}
\caption{\textit{Left}: The accuracy of the fine-tuned model and the maximum differential ability towards Chinese symbols across output representations, with respect to the number of representations ignored in the DenseNet-161 feature-extractor layer. \textit{Right}: The distribution of AUC ROC scores across output representations, with respect to the number of representations omitted from the feature extractor.}
\vspace{-0.3cm}
\label{fig:finetuning}
\end{figure}

Pre-trained ImageNet models are frequently utilized as feature extractors, where the pre-trained weights are kept fixed and only the final layer of the network is trained on a new task-specific dataset. To disable the undesired, but inherent correlations of the classes in fine-tuned networks, we propose the method that simply ignores the most sensitive representations from the feature-extractor model.
To demonstrate this, we conduct an experiment, where we employed a pre-trained DenseNet-161 model as a fixed feature-extractor and fine-tuned the last linear layer on the CalTech-256 image classification dataset \cite{griffin2007caltech} while varying the amount of the most sensitive representations omitted from the embeddings. Specifically, we ranked the representations from the DenseNet-161 feature-extractor layer based on the \textit{differentiability} towards Chinese watermarks and retrained the last linear layer while ignoring a varying amount of the most sensitive representations. To determine the effect of this procedure, we evaluated both the accuracy of each fine-tuned model, as well as the distribution of AUC ROC and differentiability scores across 256 output representations. The results of the experiment, displayed in Figure \ref{fig:finetuning}, demonstrate that by excluding 0.5\% of the most sensitive representations from the DenseNet-161 feature extractor, the dependence of the newly learned logit representations on Chinese watermarks can be significantly reduced. Furthermore, omitting up to 10\% of the most sensitive embeddings has no significant impact on the performance of the fine-tuned model while significantly suppressing the Clever-Hans effect of the new model. Additionally, it can be observed that excluding the most sensitive representations from the feature-extractor layer narrows the distribution of AUC ROC scores, making the output classes less likely to be highly differentiable towards spurious concepts.

\section{Discussion and Conclusion}

With this paper, we aim to bring awareness to the potential risks of watermarked images present in ImageNet and their impact on popular DNNs trained on this dataset. It is known that the ``carton'' class is impacted by the Chinese watermarks - however, we were able for the first time to demonstrate and identify the significant amount of other ImageNet classes, which are affected by the Chinese watermarks across popular ImageNet pre-trained models. Our results indicate that the sensitivity to watermarks is a common trait among all studied networks and this poses significant risks for transfer learning, as new models could be also vulnerable to unintended concepts. We demonstrate that by simply omitting the most watermark-sensitive representations, fine-tuned networks can suppress the reliance on the watermarks without incurring a significant decline in model performance. Overall, this study highlights the importance of paying attention to the presence of watermarks in image datasets and their impact on the performance of machine learning models.

\section*{Acknowledgements}
This work was partly funded by the German Ministry for Education and Research through the project Explaining
4.0 (ref. 01IS200551), the German Research Foundation
(ref. DFG KI-FOR 5363), the Investitionsbank Berlin through BerDiBa (grant no. 10174498), and the
European Union’s Horizon 2020 programme through iToBoS (grant no. 965221). KRM was partly funded by the German Ministry for Education and Research (under refs 01IS14013A-E, 01GQ1115, 01GQ0850, 01IS18056A, 01IS18025A and 01IS18037A) and BBDC/BZML and BIFOLD and by the Institute of Information \& Communications Technology Planning \& Evaluation (IITP) grants funded by the Korea Government (MSIT) (No. 2019-0-00079, Artificial Intelligence Graduate School Program, Korea University and No. 2022-0-00984, Development of Artificial Intelligence Technology for Personalized Plug-and-Play Explanation and Verification of Explanation).

\bibliography{main}
\bibliographystyle{iclr2023_conference}

\newpage
\appendix
\section{Appendix}
\subsection{Dataset generation}

In generating the dataset, our approach is similar to that outlined in \citet{bykov2022dora}. We implement 4 distinct scenarios, namely Chinese characters, Latin characters, Hindi characters, and Numeric watermarks. For each image in the baseline dataset, we insert a random string of 7 symbols, selected from the set of the 20 most frequently occurring characters in each language \cite{da2004corpus, trost} (for Arabic numerals we sample digits out of 10 available numbers). The watermark is placed randomly within the image, subject to the requirement of full visibility. The font size for all watermarks has been set to 30, while the image dimensions remain standard at 224 $\times$ 224 pixels.

\begin{figure}[h]
\begin{center}
\includegraphics[width=\textwidth]{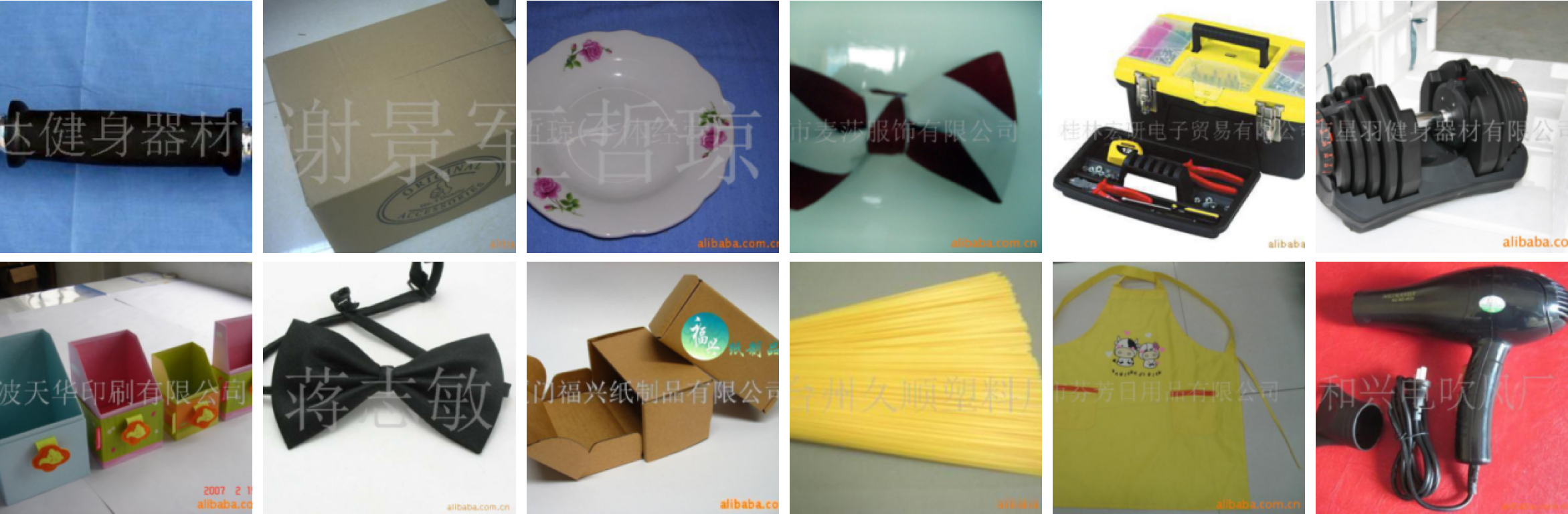}
\end{center}
\caption{Multiple images with watermarks observed in the ImageNet training dataset.}
\label{fig:appndx:wtermarks_example}
\end{figure}

\subsection{Results}

\begin{figure}[h]
\begin{center}
\includegraphics[width=\textwidth]{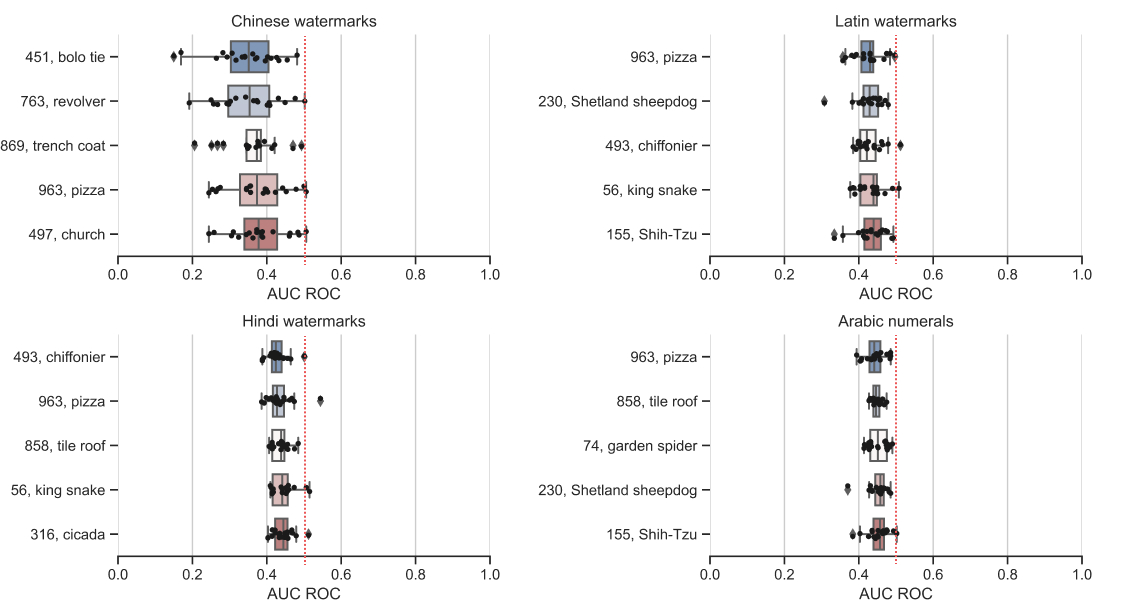}
\end{center}
\caption{Top-5 ImageNet  ranked by the lowest average AUC ROC across 20 analyzed models for 4 different scenarios.}
\label{fig:appndx:imagenet_classes_auc_roc_lowest}
\end{figure}

Figure \ref{fig:appndx:imagenet_classes_auc_roc_lowest} depicts the top-5 ImageNet classes ranked by the lowest average AUC ROC. It can be seen that, similarly to the classes with the highest AUC ROC, the ImageNet classes demonstrate a significantly better ability to differentiate between watermarked and normal images in the case of Chinese watermarks, compared to other scenarios.

\begin{figure}[h]
\begin{center}
\includegraphics[width=\textwidth]{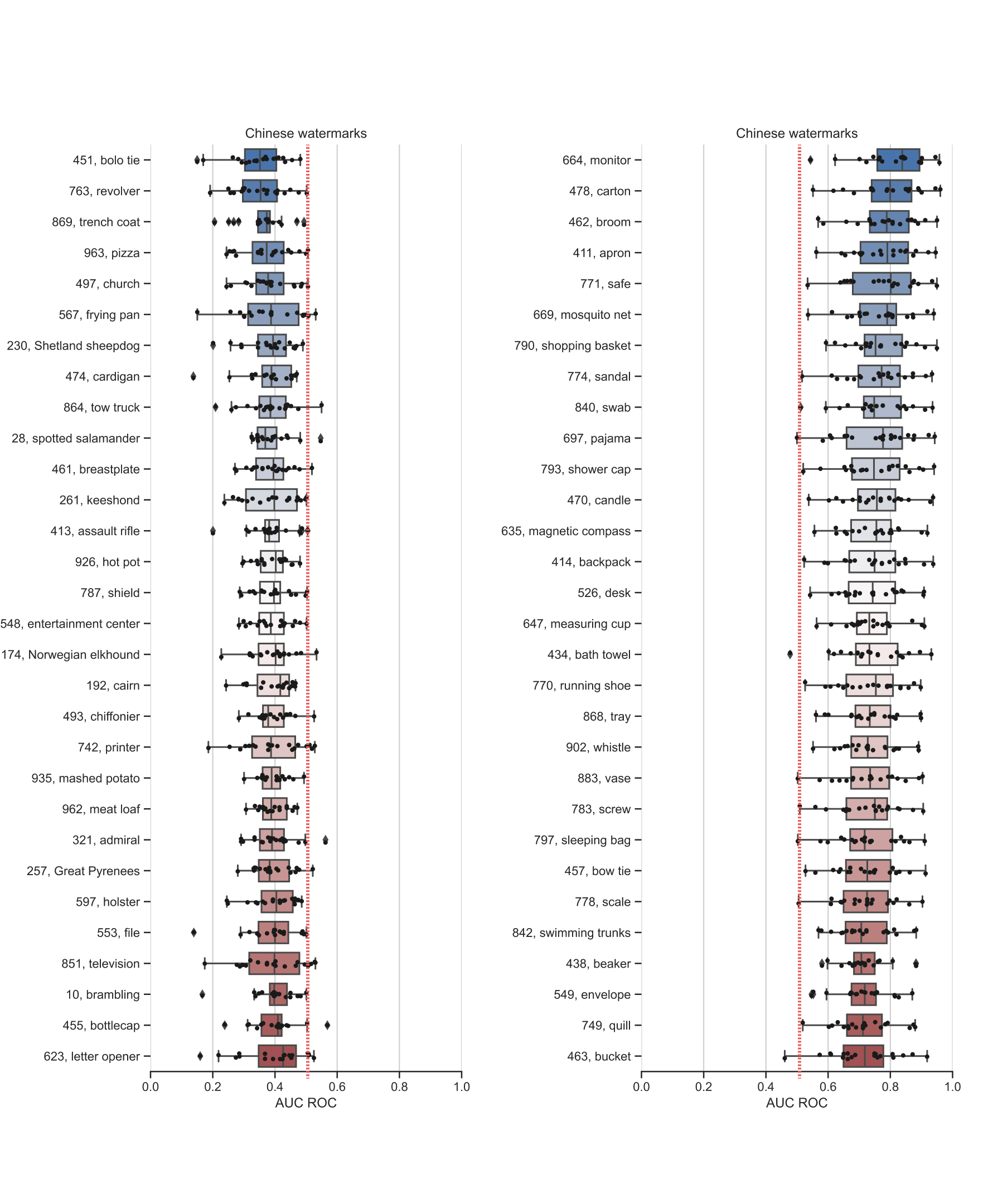}
\end{center}
\caption{\textit{Left}: The top-30 ImageNet classes ranked by the lowest average AUC ROC for the detection of Chinese symbols. \textit{Right}:the top-30 ImageNet classes ranked by the highest average AUC ROC for the detection of Chinese symbols.}
\label{fig:appndx:top30}
\end{figure}

Figure \ref{fig:appndx:top30} displays the top-30 ImageNet classes with the lowest (left) and highest (right) AUC ROC scores for the task of detection of Chinese characters.

\subsection{Ignoring sensitive embeddings during fine-tuning}

For this experiment, we utilized DenseNet-161 \cite{huang2017densely}, a well-known pre-trained model on ImageNet, to extract features from the images. The features were then subjected to an average pooling layer to yield a 2204-value embedding for each image. The embeddings were ranked based on their differentiability, i.e., their ability to distinguish between normal images and those with Chinese symbols.

To classify images on the CalTech-256 \cite{griffin2007caltech} dataset with 256 classes, we added a linear layer to the extracted features and trained the network with 10 different scenarios. In each scenario, we excluded a fraction $\alpha$ of the most differentiable embeddings from training, where $\alpha$  was varied across the  values of $0 \text{ (baseline)}, 0.005, 0.01, 0.02, 0.03, 0.05, 0.1, 0.15, 0.25, 0.5$, and trained with the same set of hyperparameters for all scenarios.
\end{document}